\documentclass{article}

\PassOptionsToPackage{numbers, compress}{natbib}
\usepackage[preprint]{neurips_2024}

\usepackage[utf8]{inputenc} % allow utf-8 input
\usepackage[T1]{fontenc}    % use 8-bit T1 fonts
\usepackage[hidelinks]{hyperref}
\usepackage{url}            % simple URL typesetting
\usepackage{booktabs}  
\usepackage{multirow} % professional-quality tables
\usepackage{amsfonts}       % blackboard math symbols
\usepackage{nicefrac}       % compact symbols for 1/2, etc.
\usepackage{microtype}      % microtypography
\usepackage{xcolor}         % colors
\usepackage{graphicx}
\usepackage{wrapfig}

\title{Scaling Diffusion Transformers to 16 Billion Parameters}
\author{%
Zhengcong Fei, Mingyuan Fan, Changqian Yu
\\
\textbf{Debang Li, Junshi Huang}\thanks{Corresponding author} \\
Kunlun Inc.\\
Beijing, China\\
{\tt\small \{feizhengcong\}@gmail.com}
}

\begin{document}

\maketitle

\begin{abstract}

In this paper, we present DiT-MoE, a sparse version of the diffusion Transformer, that is scalable and competitive with dense networks while exhibiting highly optimized inference. 
The DiT-MoE includes two simple designs: shared expert routing and expert-level balance loss, thereby capturing common knowledge and reducing redundancy among the different routed experts. When applied to conditional image generation, a deep analysis of experts specialization gains some interesting observations: (i) Expert selection shows preference with spatial position and denoising time step, while insensitive with different class-conditional information; (ii) As the MoE layers go deeper, the selection of experts gradually shifts from specific spacial position to dispersion and balance. (iii) Expert specialization tends to be more concentrated at the early time step and then gradually uniform after half. 
We attribute it to the diffusion process that first models the low-frequency spatial information and then high-frequency complex information.
Based on the above guidance, a series of DiT-MoE experimentally achieves performance on par with dense networks yet requires much less computational load during inference. 
More encouragingly, we demonstrate the potential of DiT-MoE with synthesized image data, scaling diffusion model at a 16.5B parameter that attains a new SoTA FID-50K score of 1.80 in 512$\times$512 resolution settings. 
The project page: \url{https://github.com/feizc/DiT-MoE}. 
\end{abstract}

\section{Introduction} 

Recently, diffusion models \cite{ho2020denoising,sohl2015deep,song2020score,cao2024survey} 
have emerged as powerful deep generative models in various domains, such as image \cite{dhariwal2021diffusion,ho2022cascaded,rombach2022high}, video \cite{ho2022video,mei2023vidm,singer2022make,ho2022imagen,ma2024latte}, 3D object \cite{luo2021diffusion,poole2022dreamfusion,qian2023magic123} and so on \cite{yang2023diffusion}. 
This advancement is attributed to diffusion models' ability to learn denoising tasks over diverse noise distributions, effectively transforming random noise into a target data distribution through iterative denoising processes. 
In particular, Transformer-based structure shows that increasing network capacity with additional parameters generally boosts performance \cite{chen2024pixart,peebles2023scalable,fei2024scalable,fei2024dimba}.
For example, Stable Diffusion 3 \cite{esser2024scaling} as the competitive diffusion models to date, some with over 8B parameters. However, training and serving such models is expensive \cite{patterson2021carbon}. This is partially because these deep networks are typically dense, \emph{i.e.}, every example is processed using every parameter, thereby, scale comes at a high computational cost.

Conditional computation \cite{bengio2013deep,bengio2015conditional} is a promising scaling technique, which aims to enhance model capacity while maintaining relatively constant training and inference cost by applying only a subset of parameters to each example. 
In fields of NLP, sparse mixture of experts (MoE) are becoming increasingly popular \cite{shazeer2017outrageously,dai2022stablemoe,dai2024deepseekmoe} as a practical implementation that employs a routing mechanism to control computational costs \cite{masoudnia2014mixture}. 
Moreover, the applications of MoE architectures in Transformers have yielded successful attempts at scaling language models to a substantial size with remarkable performance \cite{du2022glam,fedus2022switch,lepikhin2020gshard,zoph2022designing}.
Conventional MoE architectures in Transformers typically substitute the Feed-Forward Network (FFN) with MoE layers, each consisting of multiple experts that are structurally identical to a standard FFN.
We along with a similar sparse design and investigate its effectiveness in diffusion Transformers \cite{peebles2023scalable,ma2024sit}.

\begin{figure}[t]
  \centering
   \includegraphics[width=0.98\linewidth]{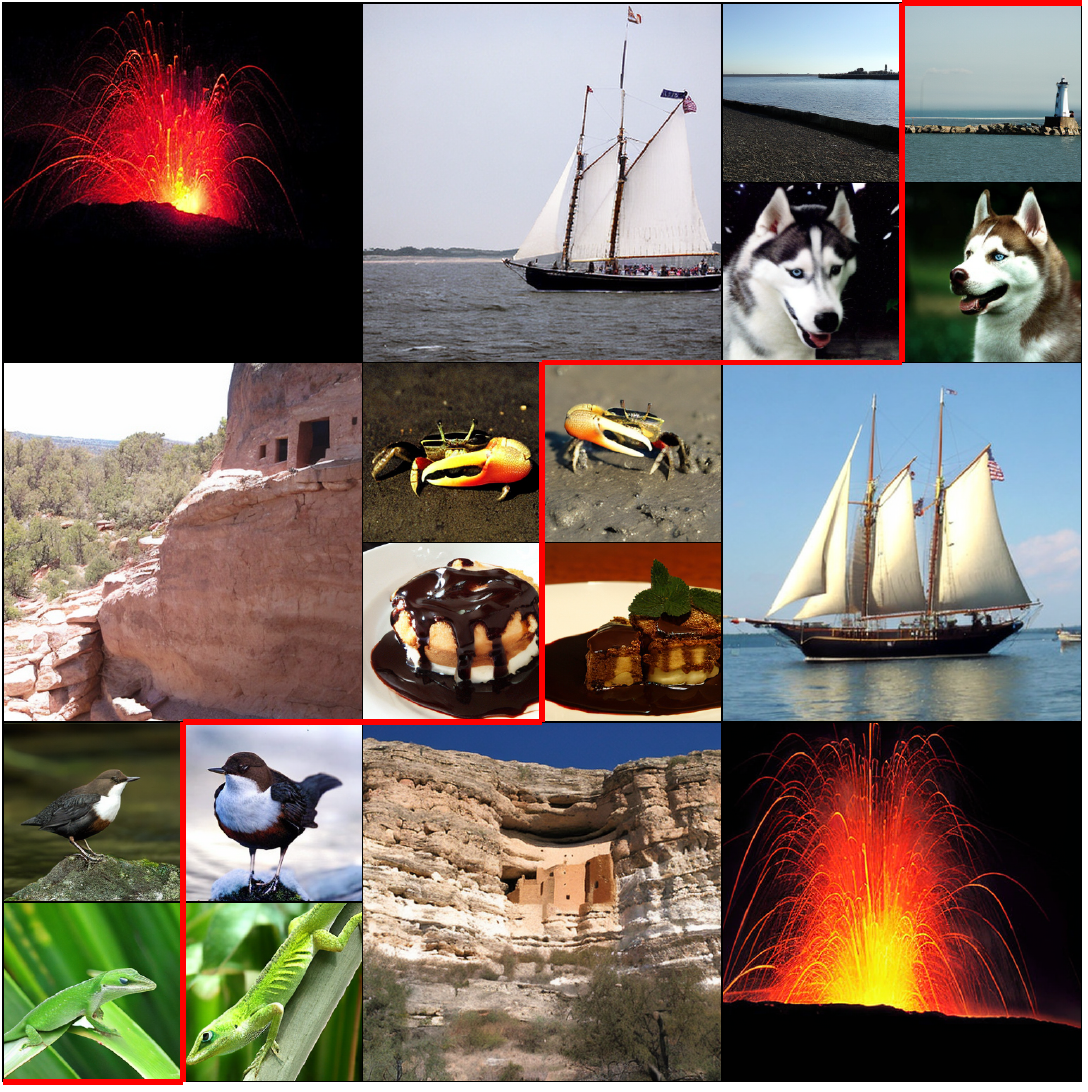}
   \caption{\textbf{DiT-MoE model achieve state-of-the-art image quality. } We show selected samples generated from our class-conditional XL/2-8E2A (left) and G/2-16E2A (right) models trained on ImageNet at 512$\times$512 and 256$\times$256 resolution, respectively. 
   }
   \label{fig:cases} 
\end{figure}

In this work, we explore conditional computation tailored specifically for Diffusion Tranformers (DiT) \cite{peebles2023scalable} at scale. We propose DiT-MoE, a sparse variant of the DiT architecture for image generation. 
The DiT-MoE replaces a subset of the dense feedforward layers in DiT with sparse MoE layers, where each token of image patch is routed to a subset of experts, \emph{i.e.}, MLP layers. 
Moreover, our architecture involves two principal designs:  shared part of experts to capture common knowledge and balance expert loss to reduce redundancy in different routed experts. 
We also provide a comprehensive analysis to demonstrate that these designs offer opportunities to train a parameter-efficient MoE diffusion model while some interesting phenomena about expert routing from different perspectives are observed.

Starting from a small-scale model, we validate the benefits of DiT-MoE architecture and present an effective recipe for the scale training of DiT-MoE. We then conduct an evaluation of class-based image generation in the ImageNet benchmarks. Experiment results indicate that DiT-MoE matches the performance of state-of-the-art dense models, while requiring less time to inference. Alternatively, DiT-MoE-S can match the cost of DiT-B while achieving better performance. 
Leveraging with additional synthesis data, we subsequently scale up the model parameters to 16.5B while only activating 3.1B parameters, which attains a new state-of-the-art FID-50K score of 1.80 in 512$\times$512 resolution.  
Our contributions can be summarized as follows: 
\begin{itemize}
    \item \textbf{MoE for diffusion transformers.} We present DiT-MoE, a sparsely-activated diffusion Transformer model for image synthesis. In between, it incorporates simple and effective designs, including shared components of experts to capture common knowledge, and an auxiliary expert-level balance loss to minimize redundancy among routed experts.
    \item  \textbf{Expert routing analysis.} We have conducted statistics on the selection of experts in different scenarios and found interesting observations about expert selection preference with spatial position and denoising time step at different MoE layers, which can effectively guide future network design and interpretability.
    \item \textbf{Model parameters at scale.} We introduce a series of DiT-MoE models and show that these models can be stably trained, and seamlessly used for efficient inference. More encouragingly, we further undertake a preliminary endeavor that DiT-MoE can performed and scale beyond 16B with well-selected synthesised data. 
    \item \textbf{Performance and inference.} We show that DiT-MoEs strongly outperform their dense counterparts on conditional image generation tasks at the ImageNet benchmark. At inference time, the DiT-MoE models can be flexible to match the performance of the largest dense model while using as little as half of the amount of computation. Finally, we publicly release the code and trained model checkpoint.
\end{itemize}

\begin{figure}[t]
  \centering
   \includegraphics[width=1.\linewidth]{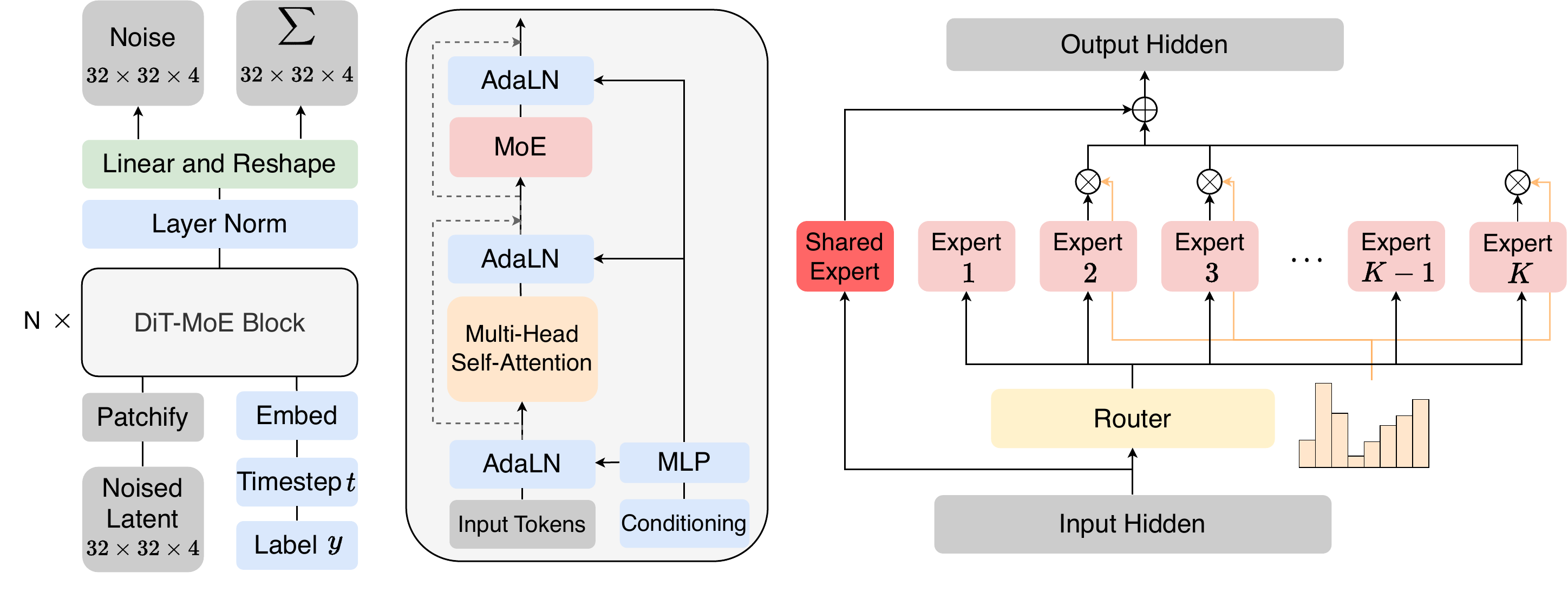}
   \caption{\textbf{Overview of the DiT-MoE architecture.} Generally, DiT-MoE is built upon the DiT and composed of MoE-inserted Transformer blocks. In between, we replace the MLP with a sparsely activated mixture of MLPs.  The right subfigure demonstrates the details of MoE layer integration of the shared expert strategy.  }
   \label{fig:framework} 
\end{figure}

\section{Methodology}

We first briefly describe diffusion models and network conditional computation with MoEs. We then present how we apply this methodology to diffusion transformers, and explain our design choices for optimizing expert routing algorithms. Finally, we provide computation analysis with different parameter scaling settings.

\subsection{Preliminaries}

\paragraph{Diffusion models.}
Diffusion models \cite{ho2020denoising,sohl2015deep} constitute a class of generative models that simulate a gradual noising and denoising process through a series of latent variables. They are characterized by a Markovian forward process and a learned reverse process. Specifically, the forward diffusion process incrementally adds noise to an input image $x_0$, transitioning it through a sequence of states $x_1, \ldots, x_T$ according to a predetermined variance schedule $\beta_1, \ldots, \beta_T$. The reverse process, learned during training, aims to recover the original data from its noised version. The forward noising process is defined as:
\begin{equation}
    q(x_t|x_0) = \mathcal{N}(\sqrt{\alpha_t}, (1 - \alpha_t)I) = \sqrt{\alpha_t} x_0 + \sqrt{(1 - \alpha_t)} \epsilon,
\end{equation}
where $\alpha_t + \beta_t = 1$ and $\epsilon \sim \mathcal{N}(0, I)$ is the Gaussian noise. 
Diffusion models are trained to estimate the reverse process, $p_\theta(x_{t-1}|x_t)$, by approximating the variational lower bound of $\int p_\theta (x_{0:T}|x_t) d(x_{0:T})$ as computed by \cite{sohl2015deep}. In practice, this reverse process is generally conditioned on the timestep $t$ and aims to either predict the noise $\epsilon$ or reconstruct the original image $x_0$. 
Formally, a noise prediction network $\epsilon_\theta ({x}_t, t)$ is incorporated by minimizing a noise prediction objective, \emph{i.e.}, $\min_\theta \mathbb{E}_{t, {x}_0, \epsilon} || \epsilon - \epsilon_\theta ({x}_t, t)||_2^2 $, where $t$ is uniformly distributed between 1 and $T$. 
To learn conditional diffusion models, \emph{e.g.}, class-conditional \cite{dhariwal2021diffusion} or text-to-image \cite{ramesh2022hierarchical,betker2023improving} models, additional condition information is integrated into the noise prediction objective as: 
\begin{equation}
    \min_\theta \mathbb{E}_{t, {x}_0, {c}, \epsilon} || \epsilon - \epsilon_\theta({x}_t, t, {c}) ||^2_2,
\end{equation}
where ${c}$ can be the condition index or its continuous embedding.

\paragraph{Conditional computation with MoEs.}

Conditional computation seeks to activate subsets of a neural network depending on the input \cite{bengio2013deep,bengio2015conditional}. 
A mixture-of-experts model exemplifies this concept by assigning different model experts to various regions of the input space \cite{jacobs1991adaptive}. 
We follow the framework of \cite{shazeer2017outrageously}, who present a mixture of experts layer in deep learning, comprising $E$ experts, defined as:
\begin{equation}
    \texttt{MoE}(x) = \sum_{i=1}^E g(x)_i e_i(x),
\end{equation}
where $x \in \mathbb{R}^{D}$ is the input to the layer, $e_i: \mathbb{R}^D \rightarrow \mathbb{R}^D$ denotes the function computed by expert $i$, and $g: \mathbb{R}^D \rightarrow \mathbb{R}^E$ is the routing function that determines the input-conditioned weights for the experts. Both $e_i$ and $g$ are parameterized by neural networks. As originally defined, this structure remains a dense network. However, if $g$ is sparse, \emph{i.e.}, restricted to assign only $k \ll E$ non-zero weights, then unused experts need not be computed. This approach enables super-linear scaling of the number of model parameters relative to the computational cost of inference and training.

\subsection{MoEs for Diffusion Transformers}

Here we explore the application of sparsity to diffuion models within the context of the Diffusion Transformers (DiT) \cite{peebles2023scalable}.
DiT has demonstrated superior scalability across various parameter settings, achieving enhanced generative performance compared to CNN-based U-Net architectures \cite{ronneberger2015u,esser2021taming} with higher training computation efficiency. 
Similar to vision transformers \cite{dosovitskiy2020image}, DiT processes images as a sequence of patches. An input image is first divided into a grid of equal-sized patches. These are linearly projected to features identical to the model's hidden dimension. After adding positional embeddings, the patch embeddings, i.e., image patch tokens, are processed by a sequence of Transformer blocks, which consists predominately of alternating self-attention and MLP layers. The standard MLPs consist of two layers and a GeLU \cite{hendrycks2016gaussian} non-linearity:
\begin{equation}
    \texttt{MLP}(x) = W_2 \sigma_{gelu} (W_1 x),
\end{equation}
For DiT-MoE, we replace a subset of these with MoE layers, where each expert is an MLP; see Figure \ref{fig:framework} for viewing. The experts share the same architecture and it follows a similar design pattern as \cite{riquelme2021scaling,dai2024deepseekmoe,du2022glam}.

On top of the generic MoE architecture, we introduce extra designs to exploit the potential of expert specialization.
As illustrated in the right subfigure \ref{fig:framework}, our architecture incorporates two principal strategies: shared expert routing and expert load balance loss. Both of these strategies are designed to optimize the level of expert specialization and introduction as below:  

\paragraph{Shared expert routing.} 
Under conventional routing strategies, tokens assigned to different experts may require access to overlapping knowledge or information. Consequently, multiple experts may converge in acquiring this shared knowledge within their respective parameters, leading to parameter redundancy.  Referring to \cite{dai2024deepseekmoe,rajbhandari2022deepspeed}, we incorporate additional $n_s$ experts to serve as shared experts. That is, regardless of the original router module, each image patch token will be deterministically assigned to these shared experts.

\paragraph{Expert-level balance loss.}
Directly learned routing strategies often encounter the issue of load imbalance, leading to significant performance defects \cite{shazeer2017sparsely}. To address this, we introduce an expert-level balance loss, calculated as follows:
\begin{equation}
    L_{balance} = \alpha \sum_{i=1}^{n} \frac{n}{K T} \sum_{t=1}^T \mathcal{I}(t, i) \frac{1}{T} \sum_{t=1}^1 \mathcal{P}(t,i),
\end{equation}
where $\alpha$ is expert-level balance factor, $T$ is the length of image patch sequence, $\mathcal{I}(t, i)$ denotes the indicator function that image token $t$ selects expert $i$ and $\mathcal{P}(t,i)$ is the probability distribution of token $t$ for expert.

\begin{table}[t]
\caption{\textbf{Scaling law model size.} The model sizes, detailed hyperparameters settings, and inference burden for MoE scaling experiments. 
}
\small
\centering
\setlength{\tabcolsep}{2.5mm}{
\begin{tabular}{lrrcccr}
\toprule
& Total param. &Activate param. & \#Blocks $L$  & Hidden dim. $D$ & \#Head $n$ &Gflops  \\ \midrule
S/2-8E2A &199M & 71M &12 & 384 &6& 15.43 \\
S/2-16E2A &369M & 71M &12 & 384 &6& 15.44 \\
B/2-8E2A &795M & 286M &12 & 768 &12& 61.68 \\
L/2-8E2A &2.8B & 1.0B &24 & 1024 &16 & 219.26 \\
XL/2-8E2A &4.1B & 1.5B &28 & 1152 &16 & 323.74 \\
G/2-16E2A &16.5B &3.1B  &40 & 1408 &16 & 690.94
\\ \bottomrule 
\end{tabular}}
\label{tab:scale}
\end{table}

\subsection{Computation Analysis}

In DiT-MoE, some of the MLPs are replaced by MoE layers, which helps increase the model capacity while keeping the activated number of parameters, and thus compute efficiency. 
Formally, the MoE modules are applied to MLPs every $e$ layer. When using MoE, there are $n$ possible experts per layer, with a router choosing the top $K$ experts and shared $n_s$ experts at each image patch token. This design allows DiT-MoE to optimize various properties by adjusting $n$, $K$, and $e$. Specifically, increasing $n$ enhances model capacity at the cost of higher memory usage, while increasing $K$ raises the number of active parameters and computational requirements. Conversely, increasing $e$ reduces model capacity but also decreases both computation and memory requirements, along with communication dependencies. 
Various configurations of DiS are delineated in Table \ref{tab:scale}. They cover a wide range of total model sizes and flop allocations, from 199M to 16.5B, thus affording comprehensive insights into the scalability performance. Aligned with \cite{peebles2023scalable}, Gflop metric is evaluated in 256$\times$256 size image generation with patch size $p=2$, checked with both \texttt{calflops} and \texttt{torchprofile} python package.
We set $e=1$ by default.  The model is named according to their configs and patch size $p$; for instance, DiT-MoE L/2-8E2A refers to the Large version config, $p=2$, $n=8$, and $K=2$.

\section{Experiments}

In this section. we begin by outlining our experimental setups in Section 3.1. Next, we present the experimental results of different DiT-MoE design spaces in Section 3.2, and provide a detailed routing analysis. Then, we provide comparative results with diffusion models in Section 3.3. Finally, we explore further scaling model with synthesized data and show some impressive cases.

\subsection{Experimental Settings}

\paragraph{Datasets.}

Following settings \cite{peebles2023scalable} for class-conditional image generation task, we utilize ImageNet \cite{deng2009imagenet} datatset at resolutions of 256$\times$256 and 512$\times$512, which comprises 1,281,167 training images across 1,000 different classes. 
The only data augmentation is horizontal flips. 
We train 500K, 1M, and 7M iterations at both resolutions, with a batch size of 1024, respectively. 
For the synthesis training data, we use open-source text-to-image models including SDXL \cite{podell2023sdxl} and SD3-Medium \cite{esser2024scaling} to create approximately 5M different 512$\times$512 images according to the given tag label. Specifically, we use the prompt template “\texttt{[image class], in a natural and realistic style.}” to create images with different seeds and filters with top CLIP similarity \cite{radford2021learning}. Finally, we construct a mixed training image set with a real-to-synthesis ratio of 1:5.

\paragraph{Implementation details. }
We use the AdamW optimizer \cite{kingma2014adam} without weight decay across all datasets, maintaining a constant learning rate of 1e-4. 
In line with \cite{peebles2023scalable}, we utilize an exponential moving average of DiT-MoE weights over training with a decay of 0.9999. All results were reported using the EMA model. Our models are trained on Nvidia A100 GPU.  
When trained on ImageNet dataset at different resolutions, we adopt classifier-free guidance \cite{ho2022classifier} following \cite{rombach2022high} and use an off-the-shelf pre-trained variational autoencoder (VAE) model \cite{kingma2013auto} from Stable Diffusion \cite{rombach2022high} available in huggingface\footnote{https://huggingface.co/stabilityai/sd-vae-ft-ema}.
The VAE encoder has a downsampling factor of 8.
We retrain diffusion hyper-parameters from \cite{peebles2023scalable}, using a $t_{max}=1000$ linear variance schedule ranging from $1\times10^{-4}$ to $2\times10^{-2}$ and parameterization of the covariance. We set the share experts number $n_s$ to 2 and the expert-level balance factor $\alpha$ to 0.005 by default.

\begin{figure}[t]
  \centering
   \includegraphics[width=1.\linewidth]{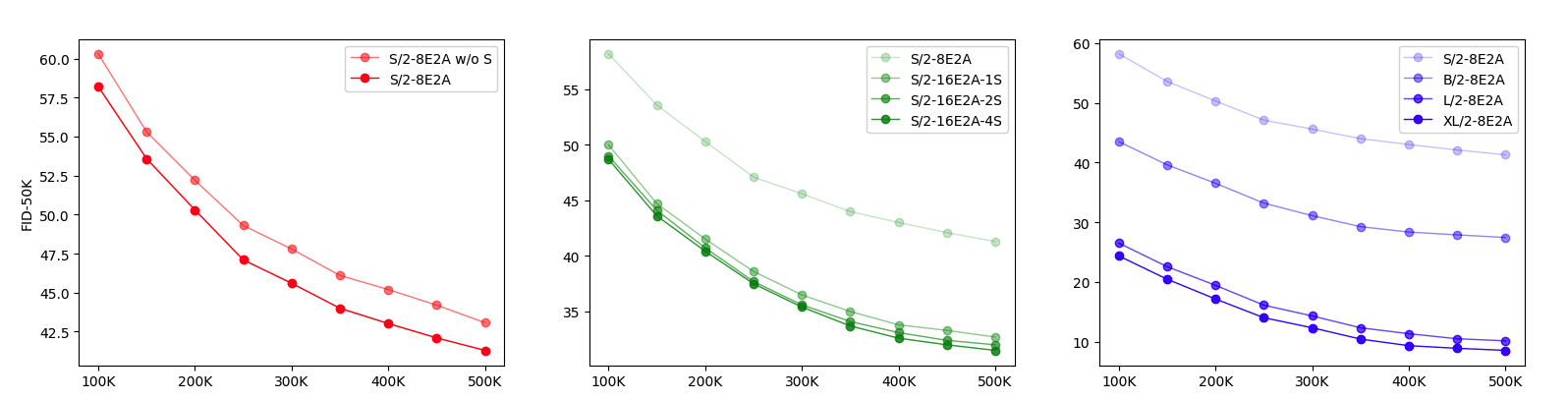}
   \caption{\textbf{Ablation experiments} on ImageNet dataset at 256$\times$256 resolution. We report FID metrics on 50K generated samples without CFG. (a) Incorporation of \textbf{shared expert routing} can accelerate the training as well as optimize generated results. 
   (b) \textbf{Number of experts} and (c) \textbf{model parameters scaling}. As we expected, increasing the expert number and the model size can consistently improve the generation performance. However, directly changing the share experts number influences the results marginally.
   }
   \label{fig:ablation} 
\end{figure}

\begin{figure}[t]
  \centering
   \includegraphics[width=1.\linewidth]{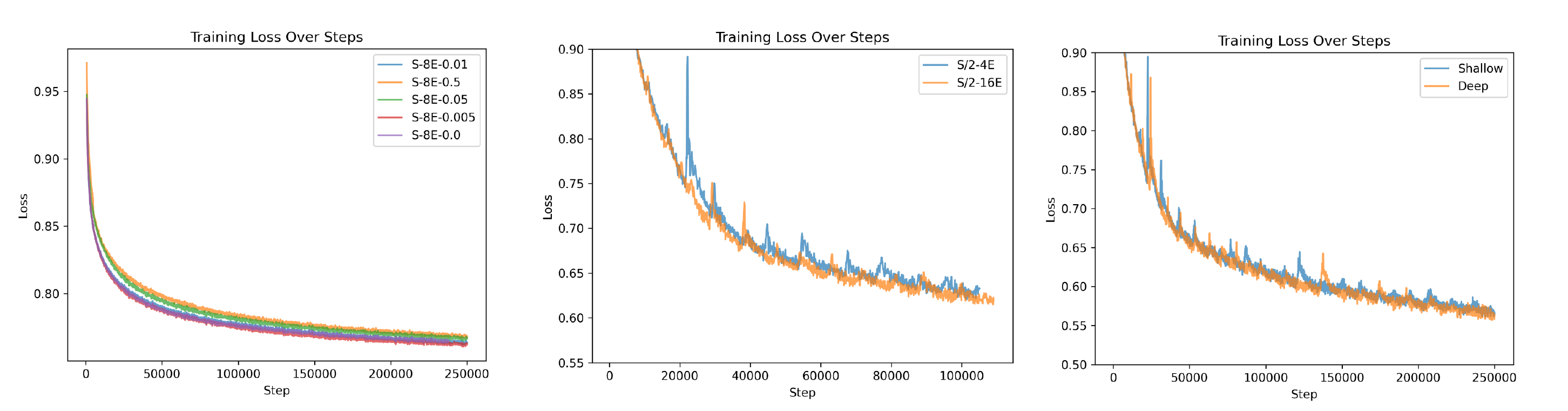}
   \caption{\textbf{Training loss curves for in small version for different elements.} (a) The effect of \textbf{expert balance loss}. We can see that 0.005 can achieve best performance. (b) The effect of \textbf{experts number} with same activate experts. We can see that increase expert number results consistently performance enhancement. (c) \textbf{The influence of MoE layer replacement.} If only half of MLP are replace with MoE layer, we can see that MoE in deep layer results in better performance, which is consisent with experted in visualization. Note that Figures are experimented in rectifide flow training.
   }
   \label{fig:abla_curve} 
\end{figure}

\paragraph{Evaluation metrics.}
We measure image generation performance with Fréchet Inception Distance (FID) \cite{heusel2017gans}, a widely adopted metric for assessing the quality of generated images.  We follow convention when comparing against prior works and report FID-50K using 250 DDPM sampling steps \cite{parmar2022aliased} following the process of \cite{dhariwal2021diffusion}. We additionally report Inception Score \cite{salimans2016improved}, sFID \cite{nash2021generating} and Precision/Recall \cite{kynkaanniemi2019improved} as secondary metrics.

\subsection{Model Design Analysis}

In this section, we ablate on the ImageNet dataset with a resolution of 256$\times$256, evaluate the FID score on 50K generated samples following \cite{bao2023all,fei2024scalable}, and identify the optimal settings.

\paragraph{Effect of shared expert routing.}
To assess the impact of the shared expert routing strategy, we conducted an ablation study by removing the shared expert while maintaining the same number of activated parameters as in the conventional expert routing approach, and trained the model from scratch. As illustrated in Figure \ref{fig:ablation} (a), the results indicate that incorporating an additional shared expert enhances performance across most steps compared to conventional expert routing. These findings support the hypothesis that the shared expert strategy facilitates better knowledge disentangling and contributes to improved MoE model performance.

\paragraph{Optimal share expert number.}
We then examine the optimal number of shared experts at scale. 
Using the small version of MoE-DiT, which comprises 16 total experts, we maintain the number of activated experts at 2 and experimented with incorporating 1, 2, and 4 shared experts. 
As depicted in Figure \ref{fig:ablation} (b), we can find that varying the ratio of shared experts to routed experts does not significantly affect performance. Considering the trade-off between memory usage and performance, we standardize the number of shared experts to 2 when scaling up DiT-MoE.

\paragraph{Effect of expert-level balance loss.}
As a import part of balance expert loading, we range from 0.5, 0.01 to 0.05, 0.005. We also dropout the expert balance loss, i.e., set to zero. As piloted in Figure \ref{fig:abla_curve} (a), as we expert, increase the expert-level balance loss leads to a performance first improve then decrease. The optimal factor is 0.005. Meantime,  we leave how to dynamically adjust the factor according to current training situation in future work.

\paragraph{Influence of increasing expert number.} 
We directly increase the expert number from 8 to 16, while keeping the number of activated parameters fixed at 2. As reported in Figure \ref{fig:ablation} (b), the adjustment leads to consistently improved generative performance, albeit with a significant increase in GPU memory consumption. 
On the other hand, the loss curve in Figure \ref{fig:abla_curve}(b) also demonstrates that incorporation of MoE in text to image tasks can achieve competitive performance and helps to faster loss convergence.

\begin{figure}[t]
  \centering
   \includegraphics[width=1\linewidth]{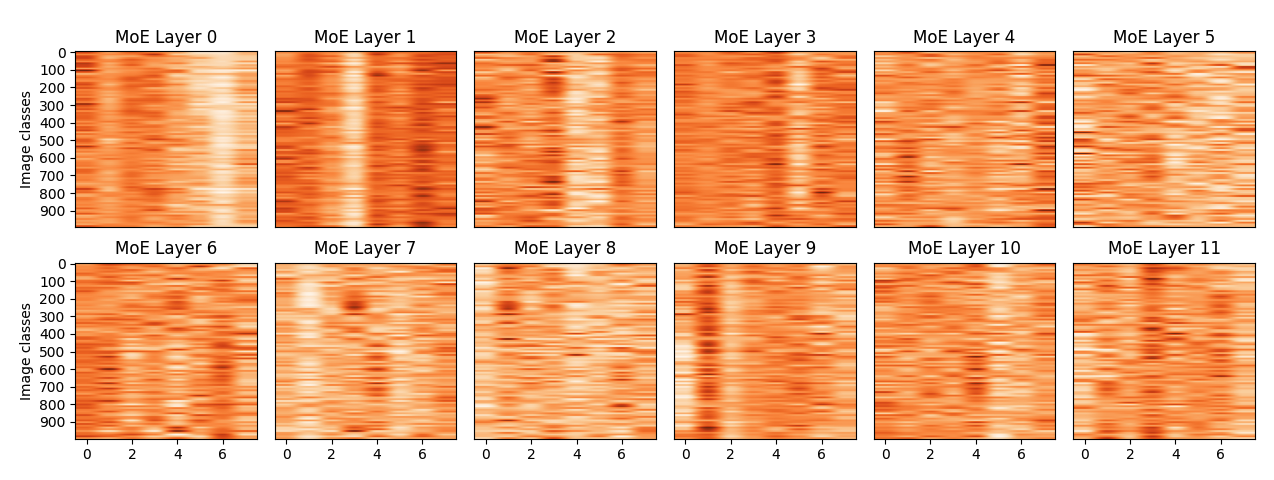}
   \caption{\textbf{Frequency for selected experts per image class.} We show the 12 MoE layers of DiT-MoE-S/2-8E2A. The $x$-axis corresponds to the 8 experts in a MoE layer. The y-axis is the 1000 ImageNet classes. For each pair (expert $e$, image class $i$), we show the average routing frequency for the patches corresponding to all generated images with class $i$ that particular expert $e$. The darker the color, the higher the frequency of selection. The larger the layer number, the deeper the MoE layers. 
   }
   \label{fig:class} 
\end{figure}

\begin{figure}[t]
  \centering
   \includegraphics[width=1\linewidth]{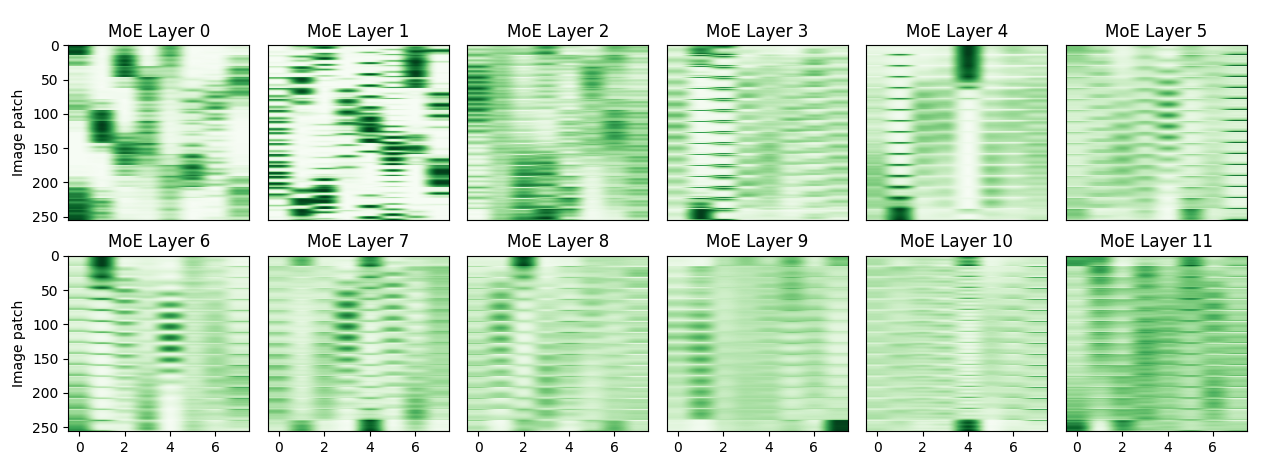}
   \caption{\textbf{Frequency for selected experts per image patch position.} We show the 12 MoE layers of DiT-MoE-S/2-8E2A. The $x$-axis corresponds to the 8 experts in a MoE layer. The $y$-axis are the 256 patches in ImageNet images with $\frac{32}{2} \times \frac{32}{2} = 256$ sequence length of patch size 2, at 256$\times$256 resolution with VAE compression 8. For each pair (expert $e$, image patch id $i$), we show the average routing frequency for all the patches with patch-id $i$ that were assigned to that particular expert $e$.
   }
   \label{fig:patch} 
\end{figure}

\paragraph{Scaling model size.} 
We also explore scaling properties of DiT-MoE by examining the effect of model depth, \emph{i.e.}, number of blocks, hidden dimension, and head number. Specifically, we train four variants of DiT-MoE model, spanning configurations from Small to XL, as detailed in Table \ref{tab:scale}, and denoted as (S, B, L, XL) for simple. 
As shown in Figure \ref{fig:ablation} (c), the performance improves as the depth increase from 12 to 28. Similarly, increasing the width from 384 to 1152 yields performance gains. 
Overall, across all configurations, impressive improvements in the FID metric are observed throughout all training stages by augmenting the depth and width of the model architecture.

\begin{figure}[t]
  \centering
   \includegraphics[width=1\linewidth]{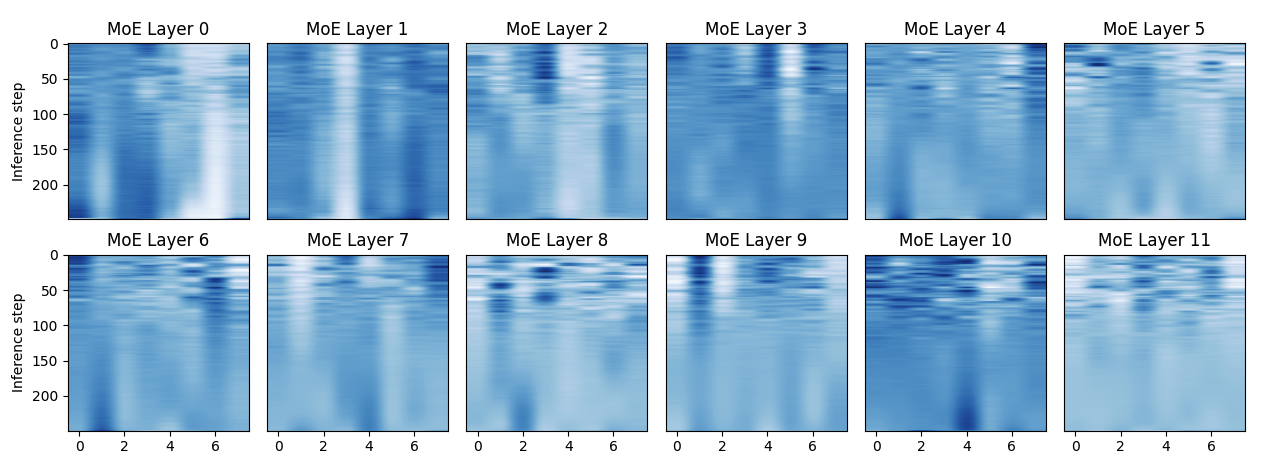} 
   \caption{\textbf{Frequency for selected experts per denoising time step.} We show the 12 MoE layers of DiT-MoE-S/2-8E2A. The x-axis corresponds to the 8 experts in a layer. The y-axis are the 250 DDPM steps for sampling the synthesis image. For each pair (expert $e$,  inference step $i$), we show the average routing frequency for all time step $i$ that were assigned to that particular expert $e$.
   }
   \label{fig:step} 
\end{figure}

\subsection{Expert Specialization Analysis}
Although large-scale MoEs have led to strong performance \cite{riquelme2021scaling,dai2024deepseekmoe}, it remains essential to explore the internal mechanisms of these complex models within the context of DiT. We posit that a thorough routing analysis can provide valuable insights for designing new algorithms. 
Specifically, we first sample 50 images for each image class with 250 DDPM steps, resulting in a total of 50K data points.
We then calculate the frequency of expert selection from three perspectives: image class, spatial position, and denoising time step. 
The visualization heat maps are presented in Figures \ref{fig:class}, \ref{fig:patch}, and \ref{fig:step}, respectively.
From our observations, several key insights emerge: 
(i) Generally, there is no obvious redundancy in the learned experts routing and each expert at a different MoE layer is routed sometimes. 
(ii) Expert selection shows a preference for spatial position and denoising step, but is less sensitive to class-conditional information, consistent with previous assumption \cite{go2023towards};
(iii) As shown in Figure \ref{fig:class}, no clear patterns or variations are evident in the expert routing mechanism for different class-conditional scenarios.
(iv) As the MoE layers become deeper, expert selection transitions from specific positional preferences to a more dispersed and balanced distribution. 
For instance, in Figure \ref{fig:patch}, the heat map of MoE layer 0 the heat map for MoE layer 0 indicates a strong correlation between image patches and spatial clustering, whereas the heat map for MoE layer 9 shows a more uniform expert selection distribution. 
(v) As in Figure \ref{fig:step}, during the early inference steps (\emph{e.g.}, steps less than 50), expert choices are more concentrated, while in later steps (\emph{e.g.}, steps greater than 100), the distribution becomes more uniform.
In summary, these findings on expert routing can effectively inform future structural designs and enhance network interpretability.

Base on the above observation, we assume that the shallow layer in DiT is sparse, and more capacity is required in deep layer. Therefore, we try to replace the dense DiT with half of MoE layer. Specifically, we replace with shallow half ($0 \sim \frac{N}{2}$) and deep half ($\frac{N}{2} \sim N$) layers with MoE layers. The results in Figure \ref{fig:abla_curve}(c) show that replacing MoE in deep layers achieve lower loss and better generative performance. 

\subsection{Compare with State-of-the-arts}

\begin{table}[t]
\centering
\small
\caption{\textbf{Benchmarking class-conditional image generation on ImageNet 256$\times$256 dataset.} We can see that DiT-MoE-XL/2 achieves state-of-the-art FID metrics towards best competitors with less inference cost.} %
    \setlength{\tabcolsep}{5.mm}{
    \begin{tabular}{lccccc}
    \toprule
    \multicolumn{6}{l}{\bf{Class-Conditional ImageNet} 256$\times$256} \\
     \midrule
    Model & FID$\downarrow$   & sFID$\downarrow$  & IS$\uparrow$     & Precision$\uparrow$ & Recall$\uparrow$ \\
      \midrule
    \multicolumn{6}{l}{ \textcolor{gray}{\emph{GAN}}} \\
    BigGAN-deep~\cite{brock2018large} & 6.95 & 7.36 & 171.4 & 0.87 & 0.28 \\
    StyleGAN-XL~\cite{sauer2022stylegan} & 2.30 & 4.02 & 265.12 & 0.78 & 0.53 \\
    \multicolumn{6}{l}{ \textcolor{gray}{\emph{Diff. based on U-Net}}}\\
    ADM~\cite{dhariwal2021diffusion} & 10.94 & 6.02 & 100.98 & 0.69 & 0.63 \\
    ADM-U & 7.49 & 5.13 & 127.49 & 0.72 & 0.63 \\
    ADM-G & 4.59 & 5.25 & 186.70 & 0.82 & 0.52 \\
    ADM-G, ADM-U & 3.94 & 6.14      & 215.84 & 0.83 & 0.53 \\
    CDM~\cite{ho2022cascaded}  & 4.88 & - & 158.71 & - & - \\
    LDM-8~\cite{rombach2022high} & 15.51 & - & 79.03 & 0.65 & 0.63 \\
    LDM-8-G & 7.76 & - & 209.52 & 0.84 & 0.35 \\
    LDM-4 & 10.56 & - & 103.49 & 0.71 & 0.62 \\
    LDM-4-G & 3.60 & - & 247.67 &  {0.87} & 0.48 \\
    VDM++ \cite{kingma2023understanding} &2.12 &-&267.70 & - & -\\
    \multicolumn{6}{l}{ \textcolor{gray}{\emph{Diff. based on Transformer}}} \\
     % {DiT-XL/2}  \cite{peebles2023scalable}             & 9.62 & 6.85 & 121.50 & 0.67 &  {0.67} \\
     % {DiT-XL/2-G} (cfg=1.25) & 3.22 & 5.28 & 201.77 & 0.76 & 0.62 \\
     U-ViT-H/2 \cite{bao2023all} &2.29 &5.68&263.88 &0.82 & 0.57 \\
     DiT-XL/2 \cite{peebles2023scalable}   &  {2.27} &  {4.60} &  {278.24} & 0.83 & 0.57 \\
     SiT-XL/2 \cite{ma2024sit} &2.06 & 4.50 &270.27 &0.82 &0.59 \\
     Large-DiT-3B \cite{gao2024lumina}  &2.10 &4.52 &304.36&0.82 &0.60 \\ 
Large-DiT-7B \cite{gao2024lumina} & 2.28 &4.35 &316.20 &0.83 &0.58 \\
LlamaGen-3B \cite{sun2024autoregressive} &2.32 & - &280.10 &0.32 &0.56\\
     DiT-MoE-XL/2-8E2A  &1.72 &4.47 & 315.73 &0.83 & 0.64 \\
    \bottomrule
    \end{tabular}}
\end{table}

We present the evaluation results of conditional image generation for various metrics compared with dense competitors in Tables 2 and 3. On the class-conditional ImageNet 256$\times$256 dataset, our DiT-MoE-XL achieves an FID score of 1.72, surpassing all previous models with different architectures. Notably, DiT-MoE-XL, which activates only 1.5 billion parameters, significantly outperforms the Transformer-based competitors Large-DiT-3B, Large-DiT-7B, and LlamaGen-3B. This demonstrates the potential of MoE in diffusion models. On the class-conditional ImageNet 512$\times$512 dataset, we observe similar advancements in nearly all evaluation metrics as expected.

\subsection{Scaling up DiT-MoE with Synthesis Data}

Building on the previous expert routing analysis, finally, we test how well DiT-MoE can scale to a very large number of parameters, while continuing to improve performance. 
For this, we expand the size of the model into giant version, detailed hyper-parameter setting listed in Table \ref{tab:scale}, and use an extensive training dataset augmented with synthesis data. 
We train a 40-block DiT-MoE model, incorporating 32 total experts with 2 activate experts, resulting in a model with 16.5B parameters while keeping a prominent inference efficient.
We successfully train DiT-MoE-G/2-16E2A, which is, as far as we are aware, the largest diffusion transformer model for class-condition image generation to date. It achieves an impressive state-of-the-art FID50K score of 1.80 at a 512$\times$512 resolution at the ImageNet benchmark. 
Figure \ref{fig:cases} showcases a selection of generated samples at different resolutions, demonstrating the high-quality image generation capacities of both DiT-MoE models.

\section{Related Works}

\paragraph{Conditional computation.}
To increase the number of model parameters without a corresponding rise in computational costs, conditional computation \cite{bengio2013deep,cho2014exponentially,davis2013low} selectively activates only relevant parts of the model based on the input, similar to decision trees \cite{loh2011classification}. This dynamic adaptation of neural networks has been applicable to various deep learning tasks \cite{bengio2013estimating,bengio2015conditional,denoyer2014deep,rosenbaum2017routing,han2021dynamic}. For instance, \cite{yang2019condconv} propose dynamically combining a set of base convolution kernels based on input image features to enhance model capacity. Additionally,  techniques in \cite{xin2020deebert,fei2022deecap,fei2019fast,fei2021partially} adjust the forward Transformer layers at the token level to expedite inference. For efficient deployment, \cite{cai2019once,yu2018slimmable} dynamically alter the neural network architecture according to specified efficiency constraints, thereby optimizing the balance between efficiency and accuracy. In a similar vein, we employ the mixture-of-experts strategy, which utilizes a gating network to dynamically route inputs to various experts.

\paragraph{Mixture of experts.}

MoEs \cite{jacobs1991adaptive,jordan1994hierarchical,chen1999improved,yuksel2012twenty} typically integrate the outputs of sub-models, or experts, through an input-dependent router, and have been successfully applied in diverse scenarios \cite{hu1997patient,gavrila2007multi,tani1999learning,zeevi1996time}. 
In the field of NLP, \cite{shazeer2017outrageously} introduced top-k gating in LSTMs, incorporating auxiliary losses to maintain expert balance \cite{hansen1999combining}. \cite{lepikhin2020gshard} extended to transformers, demonstrating substantial improvements in neural machine translation \cite{shen2019mixture}. Recent advancements in large-scale language models \cite{shazeer2017outrageously,lepikhin2020gshard,fedus2022switch} have enabled input tokens to select either all experts \cite{eigen2013learning} or a sparse mixture, facilitating the scaling of language models to trillions of parameters \cite{dai2024deepseekmoe}. \cite{gavrila2007multi} sped up pre-training with over one trillion parameters and one expert per input,  outperforming dense baseline \cite{raffel2020exploring} with transfer and distillation benefits. 
\cite{lewis2021base} alternatively employed balanced routing via a linear assignment problem. In the domain of CV, \cite{zhang2023robust,pavlitska2023sparsely} combine CNN with MoE for robust image classification.  \cite{riquelme2021scaling,ruiz2021scaling,renggli2022learning}  scale vision transformers with adaptive per-image computing,  thereby reducing the computational burden by half compared to dense competitors.

\begin{table}[t]
\centering
\small
\caption{\textbf{Benchmarking class-conditional image generation on ImageNet 512$\times$512 dataset.} DiT-MoE demonstrates a promising performance compared with dense networks for diffusion.} %
    \label{tbl:sota}
    \setlength{\tabcolsep}{5mm}{
    \begin{tabular}{lccccc}
    \toprule
    \multicolumn{6}{l}{\bf{Class-Conditional ImageNet} 512$\times$512} \\
    \midrule
    Model & FID$\downarrow$   & sFID$\downarrow$  & IS$\uparrow$     & Precision$\uparrow$ & Recall$\uparrow$ \\
     \midrule
     \multicolumn{6}{l}{ \textcolor{gray}{\emph{GAN}}} \\
    BigGAN-deep~\cite{brock2018large} & 8.43 & 8.13 & 177.90 & 0.88 & 0.29 \\
    StyleGAN-XL~\cite{sauer2022stylegan} & 2.41 & 4.06 & 267.75 & 0.77 & 0.52 \\
    \multicolumn{6}{l}{ \textcolor{gray}{\emph{Diff. based on U-Net}}}\\
    ADM~\cite{dhariwal2021diffusion} & 23.24 & 10.19 & 58.06 & 0.73 & 0.60  \\
    ADM-U & 9.96 & 5.62 & 121.78 & 0.75 &   {0.64} \\
    ADM-G & 7.72 & 6.57 & 172.71 &   {0.87} & 0.42 \\
    ADM-G, ADM-U & 3.85 & 5.86 & 221.72 & 0.84 & 0.53\\
    VDM++ \cite{kingma2023understanding} &2.65 &-&278.10 & - & -\\
    \multicolumn{6}{l}{ \textcolor{gray}{\emph{Diff. based on Transformer}}}\\
     U-ViT-H/4 \cite{bao2023all} &4.05 &6.44 &263.79&0.84 &0.48 \\
      % {DiT-XL/2-G} (cfg=1.25) & 4.64 & 5.77 & 174.77 & 0.81 & 0.57 \\
      {DiT-XL/2} \cite{peebles2023scalable} &   {3.04} &   {5.02} &   {240.82} & 0.84 & 0.54 \\ 
      Large-DiT-3B \cite{gao2024lumina} & 2.52 & 5.01 &303.70&0.82&0.57 \\
     DiT-MoE-XL/2-8E2A & 2.30 &4.82  &298.35 & 0.85 &0.57 \\
    \bottomrule
    \end{tabular}}
\end{table}

\paragraph{MoEs for diffusion models.}

While previous studies predominantly utilize a single model to tackle denoising tasks across various timesteps \cite{peebles2023scalable,fei2024dimba,podell2023sdxl,fei2024diffusion,fei2022progressive,hu2024zigma}, several investigations have explored the deployment of multiple expert models, each specializing in a distinct range of timesteps \cite{croitoru2023diffusion}. PPAP \cite{go2023towards} achieves this by training multiple classifiers on segmented timesteps, each employed for classifier guidance. e-DiffI \cite{balaji2022ediff} and ERNIE-ViLG \cite{feng2023ernie} utilize a consistent set of denoisers across these experts, whereas MEME \cite{lee2024multi} advocates for distinct architectures tailored to each timestep segment. Additionally, \cite{park2023denoising} deals with routing problem of denoising tasks across timesteps and designs a routing strategies for learning these tasks in a single model. 
These methodologies enhance generative quality while maintaining comparable inference costs, albeit at the expense of increased memory requirements. They operate on the premise that the characteristics of denoising tasks vary across timesteps. We extend it by analyzing the expert routing mechanism and demonstrating that both temporal and spatial elements without class-conditional information influence different MoE layers. 
The most similar to work is \cite{park2024switch}, which also explores experts routing, however, they focus on a form of multi-task learning for time step and not actually sparse, \emph{i.e.}, base version vs. dense version comes to 144M vs. 131M. In contrast, we delve into the time-space routing mechanism and modeling of $>$10B model size.

\section{Conclusion}

In this paper, we employ sparse conditional computation to train some of the largest diffusion transformer models, achieving efficient inference and substantial improvements in image generation tasks. Alongside DiT-MoE, we incorporate simple designs to facilitate the effective utilization of model sparsity in relation to inputs. We further provide a detailed analysis of the expert routing mechanism, demonstrating the characters of space-time preference for different MoE layers. 
This methodology aligns with recent analyses indicating that model sparsity is a highly promising strategy for reducing CO2 emissions associated with model training. 
Our work represents an initial exploration of large-scale conditional computation for diffusion models. Future extensions could involve training stable and faster, heterogeneous expert architectures and better knowledge distillation. We anticipate that the importance of sparse model scaling will continue to grow in multimodal generation.

\bibliographystyle{plainnat}
\bibliography{main}

\begin{thebibliography}{100}
\providecommand{\natexlab}[1]{#1}
\providecommand{\url}[1]{\texttt{#1}}
\expandafter\ifx\csname urlstyle\endcsname\relax
  \providecommand{\doi}[1]{doi: #1}\else
  \providecommand{\doi}{doi: \begingroup \urlstyle{rm}\Url}\fi

\bibitem[Balaji et~al.(2022)Balaji, Nah, Huang, Vahdat, Song, Zhang, Kreis, Aittala, Aila, Laine, et~al.]{balaji2022ediff}
Yogesh Balaji, Seungjun Nah, Xun Huang, Arash Vahdat, Jiaming Song, Qinsheng Zhang, Karsten Kreis, Miika Aittala, Timo Aila, Samuli Laine, et~al.
\newblock ediff-i: Text-to-image diffusion models with an ensemble of expert denoisers.
\newblock \emph{arXiv preprint arXiv:2211.01324}, 2022.

\bibitem[Bao et~al.(2023)Bao, Nie, Xue, Cao, Li, Su, and Zhu]{bao2023all}
Fan Bao, Shen Nie, Kaiwen Xue, Yue Cao, Chongxuan Li, Hang Su, and Jun Zhu.
\newblock All are worth words: A vit backbone for diffusion models.
\newblock In \emph{Proceedings of the IEEE/CVF Conference on Computer Vision and Pattern Recognition}, pages 22669--22679, 2023.

\bibitem[Bengio et~al.(2015)Bengio, Bacon, Pineau, and Precup]{bengio2015conditional}
Emmanuel Bengio, Pierre-Luc Bacon, Joelle Pineau, and Doina Precup.
\newblock Conditional computation in neural networks for faster models.
\newblock \emph{arXiv preprint arXiv:1511.06297}, 2015.

\bibitem[Bengio(2013)]{bengio2013deep}
Yoshua Bengio.
\newblock Deep learning of representations: Looking forward.
\newblock In \emph{International conference on statistical language and speech processing}, pages 1--37. Springer, 2013.

\bibitem[Bengio et~al.(2013)Bengio, L{\'e}onard, and Courville]{bengio2013estimating}
Yoshua Bengio, Nicholas L{\'e}onard, and Aaron Courville.
\newblock Estimating or propagating gradients through stochastic neurons for conditional computation.
\newblock \emph{arXiv preprint arXiv:1308.3432}, 2013.

\bibitem[Betker et~al.(2023)Betker, Goh, Jing, Brooks, Wang, Li, Ouyang, Zhuang, Lee, Guo, et~al.]{betker2023improving}
James Betker, Gabriel Goh, Li~Jing, Tim Brooks, Jianfeng Wang, Linjie Li, Long Ouyang, Juntang Zhuang, Joyce Lee, Yufei Guo, et~al.
\newblock Improving image generation with better captions.
\newblock \emph{Computer Science. https://cdn. openai. com/papers/dall-e-3. pdf}, 2\penalty0 (3):\penalty0 8, 2023.

\bibitem[Brock et~al.(2018)Brock, Donahue, and Simonyan]{brock2018large}
Andrew Brock, Jeff Donahue, and Karen Simonyan.
\newblock Large scale gan training for high fidelity natural image synthesis.
\newblock \emph{arXiv preprint arXiv:1809.11096}, 2018.

\bibitem[Cai et~al.(2019)Cai, Gan, Wang, Zhang, and Han]{cai2019once}
Han Cai, Chuang Gan, Tianzhe Wang, Zhekai Zhang, and Song Han.
\newblock Once-for-all: Train one network and specialize it for efficient deployment.
\newblock \emph{arXiv preprint arXiv:1908.09791}, 2019.

\bibitem[Cao et~al.(2024)Cao, Tan, Gao, Xu, Chen, Heng, and Li]{cao2024survey}
Hanqun Cao, Cheng Tan, Zhangyang Gao, Yilun Xu, Guangyong Chen, Pheng-Ann Heng, and Stan~Z Li.
\newblock A survey on generative diffusion models.
\newblock \emph{IEEE Transactions on Knowledge and Data Engineering}, 2024.

\bibitem[Chen et~al.(2024)Chen, Ge, Xie, Wu, Yao, Ren, Wang, Luo, Lu, and Li]{chen2024pixart}
Junsong Chen, Chongjian Ge, Enze Xie, Yue Wu, Lewei Yao, Xiaozhe Ren, Zhongdao Wang, Ping Luo, Huchuan Lu, and Zhenguo Li.
\newblock Pixart-$\backslash$sigma: Weak-to-strong training of diffusion transformer for 4k text-to-image generation.
\newblock \emph{arXiv preprint arXiv:2403.04692}, 2024.

\bibitem[Chen et~al.(1999)Chen, Xu, and Chi]{chen1999improved}
Ke~Chen, Lei Xu, and Huisheng Chi.
\newblock Improved learning algorithms for mixture of experts in multiclass classification.
\newblock \emph{Neural networks}, 12\penalty0 (9):\penalty0 1229--1252, 1999.

\bibitem[Cho and Bengio(2014)]{cho2014exponentially}
Kyunghyun Cho and Yoshua Bengio.
\newblock Exponentially increasing the capacity-to-computation ratio for conditional computation in deep learning.
\newblock \emph{arXiv preprint arXiv:1406.7362}, 2014.

\bibitem[Croitoru et~al.(2023)Croitoru, Hondru, Ionescu, and Shah]{croitoru2023diffusion}
Florinel-Alin Croitoru, Vlad Hondru, Radu~Tudor Ionescu, and Mubarak Shah.
\newblock Diffusion models in vision: A survey.
\newblock \emph{IEEE Transactions on Pattern Analysis and Machine Intelligence}, 2023.

\bibitem[Dai et~al.(2022)Dai, Dong, Ma, Zheng, Sui, Chang, and Wei]{dai2022stablemoe}
Damai Dai, Li~Dong, Shuming Ma, Bo~Zheng, Zhifang Sui, Baobao Chang, and Furu Wei.
\newblock Stablemoe: Stable routing strategy for mixture of experts.
\newblock \emph{arXiv preprint arXiv:2204.08396}, 2022.

\bibitem[Dai et~al.(2024)Dai, Deng, Zhao, Xu, Gao, Chen, Li, Zeng, Yu, Wu, et~al.]{dai2024deepseekmoe}
Damai Dai, Chengqi Deng, Chenggang Zhao, RX~Xu, Huazuo Gao, Deli Chen, Jiashi Li, Wangding Zeng, Xingkai Yu, Y~Wu, et~al.
\newblock Deepseekmoe: Towards ultimate expert specialization in mixture-of-experts language models.
\newblock \emph{arXiv preprint arXiv:2401.06066}, 2024.

\bibitem[Davis and Arel(2013)]{davis2013low}
Andrew Davis and Itamar Arel.
\newblock Low-rank approximations for conditional feedforward computation in deep neural networks.
\newblock \emph{arXiv preprint arXiv:1312.4461}, 2013.

\bibitem[Deng et~al.(2009)Deng, Dong, Socher, Li, Li, and Fei-Fei]{deng2009imagenet}
Jia Deng, Wei Dong, Richard Socher, Li-Jia Li, Kai Li, and Li~Fei-Fei.
\newblock Imagenet: A large-scale hierarchical image database.
\newblock In \emph{2009 IEEE conference on computer vision and pattern recognition}, pages 248--255. Ieee, 2009.

\bibitem[Denoyer and Gallinari(2014)]{denoyer2014deep}
Ludovic Denoyer and Patrick Gallinari.
\newblock Deep sequential neural network.
\newblock \emph{arXiv preprint arXiv:1410.0510}, 2014.

\bibitem[Dhariwal and Nichol(2021)]{dhariwal2021diffusion}
Prafulla Dhariwal and Alexander Nichol.
\newblock Diffusion models beat gans on image synthesis.
\newblock \emph{Advances in neural information processing systems}, 34:\penalty0 8780--8794, 2021.

\bibitem[Dosovitskiy et~al.(2020)Dosovitskiy, Beyer, Kolesnikov, Weissenborn, Zhai, Unterthiner, Dehghani, Minderer, Heigold, Gelly, et~al.]{dosovitskiy2020image}
Alexey Dosovitskiy, Lucas Beyer, Alexander Kolesnikov, Dirk Weissenborn, Xiaohua Zhai, Thomas Unterthiner, Mostafa Dehghani, Matthias Minderer, Georg Heigold, Sylvain Gelly, et~al.
\newblock An image is worth 16x16 words: Transformers for image recognition at scale.
\newblock \emph{arXiv preprint arXiv:2010.11929}, 2020.

\bibitem[Du et~al.(2022)Du, Huang, Dai, Tong, Lepikhin, Xu, Krikun, Zhou, Yu, Firat, et~al.]{du2022glam}
Nan Du, Yanping Huang, Andrew~M Dai, Simon Tong, Dmitry Lepikhin, Yuanzhong Xu, Maxim Krikun, Yanqi Zhou, Adams~Wei Yu, Orhan Firat, et~al.
\newblock Glam: Efficient scaling of language models with mixture-of-experts.
\newblock In \emph{International Conference on Machine Learning}, pages 5547--5569. PMLR, 2022.

\bibitem[Eigen et~al.(2013)Eigen, Ranzato, and Sutskever]{eigen2013learning}
David Eigen, Marc'Aurelio Ranzato, and Ilya Sutskever.
\newblock Learning factored representations in a deep mixture of experts.
\newblock \emph{arXiv preprint arXiv:1312.4314}, 2013.

\bibitem[Esser et~al.(2021)Esser, Rombach, and Ommer]{esser2021taming}
Patrick Esser, Robin Rombach, and Bjorn Ommer.
\newblock Taming transformers for high-resolution image synthesis.
\newblock In \emph{Proceedings of the IEEE/CVF conference on computer vision and pattern recognition}, pages 12873--12883, 2021.

\bibitem[Esser et~al.(2024)Esser, Kulal, Blattmann, Entezari, M{\"u}ller, Saini, Levi, Lorenz, Sauer, Boesel, et~al.]{esser2024scaling}
Patrick Esser, Sumith Kulal, Andreas Blattmann, Rahim Entezari, Jonas M{\"u}ller, Harry Saini, Yam Levi, Dominik Lorenz, Axel Sauer, Frederic Boesel, et~al.
\newblock Scaling rectified flow transformers for high-resolution image synthesis.
\newblock In \emph{Forty-first International Conference on Machine Learning}, 2024.

\bibitem[Fedus et~al.(2022)Fedus, Zoph, and Shazeer]{fedus2022switch}
William Fedus, Barret Zoph, and Noam Shazeer.
\newblock Switch transformers: Scaling to trillion parameter models with simple and efficient sparsity.
\newblock \emph{Journal of Machine Learning Research}, 23\penalty0 (120):\penalty0 1--39, 2022.

\bibitem[Fei(2019)]{fei2019fast}
Zheng-cong Fei.
\newblock Fast image caption generation with position alignment.
\newblock \emph{arXiv preprint arXiv:1912.06365}, 2019.

\bibitem[Fei(2021)]{fei2021partially}
Zhengcong Fei.
\newblock Partially non-autoregressive image captioning.
\newblock In \emph{Proceedings of the AAAI Conference on Artificial Intelligence}, volume~35, pages 1309--1316, 2021.

\bibitem[Fei et~al.(2022{\natexlab{a}})Fei, Fan, Zhu, and Huang]{fei2022progressive}
Zhengcong Fei, Mingyuan Fan, Li~Zhu, and Junshi Huang.
\newblock Progressive text-to-image generation.
\newblock \emph{arXiv preprint arXiv:2210.02291}, 2022{\natexlab{a}}.

\bibitem[Fei et~al.(2022{\natexlab{b}})Fei, Yan, Wang, and Tian]{fei2022deecap}
Zhengcong Fei, Xu~Yan, Shuhui Wang, and Qi~Tian.
\newblock Deecap: Dynamic early exiting for efficient image captioning.
\newblock In \emph{Proceedings of the IEEE/CVF Conference on Computer Vision and Pattern Recognition}, pages 12216--12226, 2022{\natexlab{b}}.

\bibitem[Fei et~al.(2024{\natexlab{a}})Fei, Fan, Yu, and Huang]{fei2024scalable}
Zhengcong Fei, Mingyuan Fan, Changqian Yu, and Junshi Huang.
\newblock Scalable diffusion models with state space backbone.
\newblock \emph{arXiv preprint arXiv:2402.05608}, 2024{\natexlab{a}}.

\bibitem[Fei et~al.(2024{\natexlab{b}})Fei, Fan, Yu, Li, and Huang]{fei2024diffusion}
Zhengcong Fei, Mingyuan Fan, Changqian Yu, Debang Li, and Junshi Huang.
\newblock Diffusion-rwkv: Scaling rwkv-like architectures for diffusion models.
\newblock \emph{arXiv preprint arXiv:2404.04478}, 2024{\natexlab{b}}.

\bibitem[Fei et~al.(2024{\natexlab{c}})Fei, Fan, Yu, Li, Zhang, and Huang]{fei2024dimba}
Zhengcong Fei, Mingyuan Fan, Changqian Yu, Debang Li, Youqiang Zhang, and Junshi Huang.
\newblock Dimba: Transformer-mamba diffusion models.
\newblock \emph{arXiv preprint arXiv:2406.01159}, 2024{\natexlab{c}}.

\bibitem[Feng et~al.(2023)Feng, Zhang, Yu, Fang, Li, Chen, Lu, Liu, Yin, Feng, et~al.]{feng2023ernie}
Zhida Feng, Zhenyu Zhang, Xintong Yu, Yewei Fang, Lanxin Li, Xuyi Chen, Yuxiang Lu, Jiaxiang Liu, Weichong Yin, Shikun Feng, et~al.
\newblock Ernie-vilg 2.0: Improving text-to-image diffusion model with knowledge-enhanced mixture-of-denoising-experts.
\newblock In \emph{Proceedings of the IEEE/CVF Conference on Computer Vision and Pattern Recognition}, pages 10135--10145, 2023.

\bibitem[Gao et~al.(2024)Gao, Zhuo, Lin, Liu, Chen, Du, Xie, Luo, Qiu, Zhang, et~al.]{gao2024lumina}
Peng Gao, Le~Zhuo, Ziyi Lin, Chris Liu, Junsong Chen, Ruoyi Du, Enze Xie, Xu~Luo, Longtian Qiu, Yuhang Zhang, et~al.
\newblock Lumina-t2x: Transforming text into any modality, resolution, and duration via flow-based large diffusion transformers.
\newblock \emph{arXiv preprint arXiv:2405.05945}, 2024.

\bibitem[Gavrila and Munder(2007)]{gavrila2007multi}
Dariu~M Gavrila and Stefan Munder.
\newblock Multi-cue pedestrian detection and tracking from a moving vehicle.
\newblock \emph{International journal of computer vision}, 73:\penalty0 41--59, 2007.

\bibitem[Go et~al.(2023)Go, Lee, Kim, Lee, Jeong, Lee, and Choi]{go2023towards}
Hyojun Go, Yunsung Lee, Jin-Young Kim, Seunghyun Lee, Myeongho Jeong, Hyun~Seung Lee, and Seungtaek Choi.
\newblock Towards practical plug-and-play diffusion models.
\newblock In \emph{Proceedings of the IEEE/CVF conference on computer vision and pattern recognition}, pages 1962--1971, 2023.

\bibitem[Han et~al.(2021)Han, Huang, Song, Yang, Wang, and Wang]{han2021dynamic}
Yizeng Han, Gao Huang, Shiji Song, Le~Yang, Honghui Wang, and Yulin Wang.
\newblock Dynamic neural networks: A survey.
\newblock \emph{IEEE Transactions on Pattern Analysis and Machine Intelligence}, 44\penalty0 (11):\penalty0 7436--7456, 2021.

\bibitem[Hansen(1999)]{hansen1999combining}
Jakob~Vogdrup Hansen.
\newblock Combining predictors: comparison of five meta machine learning methods.
\newblock \emph{Information Sciences}, 119\penalty0 (1-2):\penalty0 91--105, 1999.

\bibitem[Hendrycks and Gimpel(2016)]{hendrycks2016gaussian}
Dan Hendrycks and Kevin Gimpel.
\newblock Gaussian error linear units (gelus).
\newblock \emph{arXiv preprint arXiv:1606.08415}, 2016.

\bibitem[Heusel et~al.(2017)Heusel, Ramsauer, Unterthiner, Nessler, and Hochreiter]{heusel2017gans}
Martin Heusel, Hubert Ramsauer, Thomas Unterthiner, Bernhard Nessler, and Sepp Hochreiter.
\newblock Gans trained by a two time-scale update rule converge to a local nash equilibrium.
\newblock \emph{Advances in neural information processing systems}, 30, 2017.

\bibitem[Ho and Salimans(2022)]{ho2022classifier}
Jonathan Ho and Tim Salimans.
\newblock Classifier-free diffusion guidance.
\newblock \emph{arXiv preprint arXiv:2207.12598}, 2022.

\bibitem[Ho et~al.(2020)Ho, Jain, and Abbeel]{ho2020denoising}
Jonathan Ho, Ajay Jain, and Pieter Abbeel.
\newblock Denoising diffusion probabilistic models.
\newblock \emph{Advances in neural information processing systems}, 33:\penalty0 6840--6851, 2020.

\bibitem[Ho et~al.(2022{\natexlab{a}})Ho, Chan, Saharia, Whang, Gao, Gritsenko, Kingma, Poole, Norouzi, Fleet, et~al.]{ho2022imagen}
Jonathan Ho, William Chan, Chitwan Saharia, Jay Whang, Ruiqi Gao, Alexey Gritsenko, Diederik~P Kingma, Ben Poole, Mohammad Norouzi, David~J Fleet, et~al.
\newblock Imagen video: High definition video generation with diffusion models.
\newblock \emph{arXiv preprint arXiv:2210.02303}, 2022{\natexlab{a}}.

\bibitem[Ho et~al.(2022{\natexlab{b}})Ho, Saharia, Chan, Fleet, Norouzi, and Salimans]{ho2022cascaded}
Jonathan Ho, Chitwan Saharia, William Chan, David~J Fleet, Mohammad Norouzi, and Tim Salimans.
\newblock Cascaded diffusion models for high fidelity image generation.
\newblock \emph{The Journal of Machine Learning Research}, 23\penalty0 (1):\penalty0 2249--2281, 2022{\natexlab{b}}.

\bibitem[Ho et~al.(2022{\natexlab{c}})Ho, Salimans, Gritsenko, Chan, Norouzi, and Fleet]{ho2022video}
Jonathan Ho, Tim Salimans, Alexey Gritsenko, William Chan, Mohammad Norouzi, and David~J Fleet.
\newblock Video diffusion models.
\newblock \emph{Advances in Neural Information Processing Systems}, 35:\penalty0 8633--8646, 2022{\natexlab{c}}.

\bibitem[Hu et~al.(2024)Hu, Baumann, Gui, Grebenkova, Ma, Fischer, and Ommer]{hu2024zigma}
Vincent~Tao Hu, Stefan~Andreas Baumann, Ming Gui, Olga Grebenkova, Pingchuan Ma, Johannes~S Fischer, and Bj{\"o}rn Ommer.
\newblock Zigma: A dit-style zigzag mamba diffusion model.
\newblock \emph{arXiv preprint arXiv:2403.13802}, 2024.

\bibitem[Hu et~al.(1997)Hu, Palreddy, and Tompkins]{hu1997patient}
Yu~Hen Hu, Surekha Palreddy, and Willis~J Tompkins.
\newblock A patient-adaptable ecg beat classifier using a mixture of experts approach.
\newblock \emph{IEEE transactions on biomedical engineering}, 44\penalty0 (9):\penalty0 891--900, 1997.

\bibitem[Jacobs et~al.(1991)Jacobs, Jordan, Nowlan, and Hinton]{jacobs1991adaptive}
Robert~A Jacobs, Michael~I Jordan, Steven~J Nowlan, and Geoffrey~E Hinton.
\newblock Adaptive mixtures of local experts.
\newblock \emph{Neural computation}, 3\penalty0 (1):\penalty0 79--87, 1991.

\bibitem[Jordan and Jacobs(1994)]{jordan1994hierarchical}
Michael~I Jordan and Robert~A Jacobs.
\newblock Hierarchical mixtures of experts and the em algorithm.
\newblock \emph{Neural computation}, 6\penalty0 (2):\penalty0 181--214, 1994.

\bibitem[Kingma and Ba(2014)]{kingma2014adam}
Diederik~P Kingma and Jimmy Ba.
\newblock Adam: A method for stochastic optimization.
\newblock \emph{arXiv preprint arXiv:1412.6980}, 2014.

\bibitem[Kingma and Gao(2023)]{kingma2023understanding}
Diederik~P Kingma and Ruiqi Gao.
\newblock Understanding diffusion objectives as the elbo with simple data augmentation.
\newblock In \emph{Thirty-seventh Conference on Neural Information Processing Systems}, 2023.

\bibitem[Kingma and Welling(2013)]{kingma2013auto}
Diederik~P Kingma and Max Welling.
\newblock Auto-encoding variational bayes.
\newblock \emph{arXiv preprint arXiv:1312.6114}, 2013.

\bibitem[Kynk{\"a}{\"a}nniemi et~al.(2019)Kynk{\"a}{\"a}nniemi, Karras, Laine, Lehtinen, and Aila]{kynkaanniemi2019improved}
Tuomas Kynk{\"a}{\"a}nniemi, Tero Karras, Samuli Laine, Jaakko Lehtinen, and Timo Aila.
\newblock Improved precision and recall metric for assessing generative models.
\newblock \emph{Advances in Neural Information Processing Systems}, 32, 2019.

\bibitem[Lee et~al.(2024)Lee, Kim, Go, Jeong, Oh, and Choi]{lee2024multi}
Yunsung Lee, JinYoung Kim, Hyojun Go, Myeongho Jeong, Shinhyeok Oh, and Seungtaek Choi.
\newblock Multi-architecture multi-expert diffusion models.
\newblock In \emph{Proceedings of the AAAI Conference on Artificial Intelligence}, volume~38, pages 13427--13436, 2024.

\bibitem[Lepikhin et~al.(2020)Lepikhin, Lee, Xu, Chen, Firat, Huang, Krikun, Shazeer, and Chen]{lepikhin2020gshard}
Dmitry Lepikhin, HyoukJoong Lee, Yuanzhong Xu, Dehao Chen, Orhan Firat, Yanping Huang, Maxim Krikun, Noam Shazeer, and Zhifeng Chen.
\newblock Gshard: Scaling giant models with conditional computation and automatic sharding.
\newblock \emph{arXiv preprint arXiv:2006.16668}, 2020.

\bibitem[Lewis et~al.(2021)Lewis, Bhosale, Dettmers, Goyal, and Zettlemoyer]{lewis2021base}
Mike Lewis, Shruti Bhosale, Tim Dettmers, Naman Goyal, and Luke Zettlemoyer.
\newblock Base layers: Simplifying training of large, sparse models.
\newblock In \emph{International Conference on Machine Learning}, pages 6265--6274. PMLR, 2021.

\bibitem[Loh(2011)]{loh2011classification}
Wei-Yin Loh.
\newblock Classification and regression trees.
\newblock \emph{Wiley interdisciplinary reviews: data mining and knowledge discovery}, 1\penalty0 (1):\penalty0 14--23, 2011.

\bibitem[Luo and Hu(2021)]{luo2021diffusion}
Shitong Luo and Wei Hu.
\newblock Diffusion probabilistic models for 3d point cloud generation.
\newblock In \emph{Proceedings of the IEEE/CVF conference on computer vision and pattern recognition}, pages 2837--2845, 2021.

\bibitem[Ma et~al.(2024{\natexlab{a}})Ma, Goldstein, Albergo, Boffi, Vanden-Eijnden, and Xie]{ma2024sit}
Nanye Ma, Mark Goldstein, Michael~S Albergo, Nicholas~M Boffi, Eric Vanden-Eijnden, and Saining Xie.
\newblock Sit: Exploring flow and diffusion-based generative models with scalable interpolant transformers.
\newblock \emph{arXiv preprint arXiv:2401.08740}, 2024{\natexlab{a}}.

\bibitem[Ma et~al.(2024{\natexlab{b}})Ma, Wang, Jia, Chen, Liu, Li, Chen, and Qiao]{ma2024latte}
Xin Ma, Yaohui Wang, Gengyun Jia, Xinyuan Chen, Ziwei Liu, Yuan-Fang Li, Cunjian Chen, and Yu~Qiao.
\newblock Latte: Latent diffusion transformer for video generation.
\newblock \emph{arXiv preprint arXiv:2401.03048}, 2024{\natexlab{b}}.

\bibitem[Masoudnia and Ebrahimpour(2014)]{masoudnia2014mixture}
Saeed Masoudnia and Reza Ebrahimpour.
\newblock Mixture of experts: a literature survey.
\newblock \emph{Artificial Intelligence Review}, 42:\penalty0 275--293, 2014.

\bibitem[Mei and Patel(2023)]{mei2023vidm}
Kangfu Mei and Vishal Patel.
\newblock Vidm: Video implicit diffusion models.
\newblock In \emph{Proceedings of the AAAI Conference on Artificial Intelligence}, volume~37, pages 9117--9125, 2023.

\bibitem[Nash et~al.(2021)Nash, Menick, Dieleman, and Battaglia]{nash2021generating}
Charlie Nash, Jacob Menick, Sander Dieleman, and Peter~W Battaglia.
\newblock Generating images with sparse representations.
\newblock \emph{arXiv preprint arXiv:2103.03841}, 2021.

\bibitem[Park et~al.(2023)Park, Woo, Go, Kim, and Kim]{park2023denoising}
Byeongjun Park, Sangmin Woo, Hyojun Go, Jin-Young Kim, and Changick Kim.
\newblock Denoising task routing for diffusion models.
\newblock \emph{arXiv preprint arXiv:2310.07138}, 2023.

\bibitem[Park et~al.(2024)Park, Go, Kim, Woo, Ham, and Kim]{park2024switch}
Byeongjun Park, Hyojun Go, Jin-Young Kim, Sangmin Woo, Seokil Ham, and Changick Kim.
\newblock Switch diffusion transformer: Synergizing denoising tasks with sparse mixture-of-experts.
\newblock \emph{arXiv preprint arXiv:2403.09176}, 2024.

\bibitem[Parmar et~al.(2022)Parmar, Zhang, and Zhu]{parmar2022aliased}
Gaurav Parmar, Richard Zhang, and Jun-Yan Zhu.
\newblock On aliased resizing and surprising subtleties in gan evaluation.
\newblock In \emph{Proceedings of the IEEE/CVF Conference on Computer Vision and Pattern Recognition}, pages 11410--11420, 2022.

\bibitem[Patterson et~al.(2021)Patterson, Gonzalez, Le, Liang, Munguia, Rothchild, So, Texier, and Dean]{patterson2021carbon}
David Patterson, Joseph Gonzalez, Quoc Le, Chen Liang, Lluis-Miquel Munguia, Daniel Rothchild, David So, Maud Texier, and Jeff Dean.
\newblock Carbon emissions and large neural network training.
\newblock \emph{arXiv preprint arXiv:2104.10350}, 2021.

\bibitem[Pavlitska et~al.(2023)Pavlitska, Hubschneider, Struppek, and Z{\"o}llner]{pavlitska2023sparsely}
Svetlana Pavlitska, Christian Hubschneider, Lukas Struppek, and J~Marius Z{\"o}llner.
\newblock Sparsely-gated mixture-of-expert layers for cnn interpretability.
\newblock In \emph{2023 International Joint Conference on Neural Networks (IJCNN)}, pages 1--10. IEEE, 2023.

\bibitem[Peebles and Xie(2023)]{peebles2023scalable}
William Peebles and Saining Xie.
\newblock Scalable diffusion models with transformers.
\newblock In \emph{Proceedings of the IEEE/CVF International Conference on Computer Vision}, pages 4195--4205, 2023.

\bibitem[Podell et~al.(2023)Podell, English, Lacey, Blattmann, Dockhorn, M{\"u}ller, Penna, and Rombach]{podell2023sdxl}
Dustin Podell, Zion English, Kyle Lacey, Andreas Blattmann, Tim Dockhorn, Jonas M{\"u}ller, Joe Penna, and Robin Rombach.
\newblock Sdxl: Improving latent diffusion models for high-resolution image synthesis.
\newblock \emph{arXiv preprint arXiv:2307.01952}, 2023.

\bibitem[Poole et~al.(2022)Poole, Jain, Barron, and Mildenhall]{poole2022dreamfusion}
Ben Poole, Ajay Jain, Jonathan~T Barron, and Ben Mildenhall.
\newblock Dreamfusion: Text-to-3d using 2d diffusion.
\newblock \emph{arXiv preprint arXiv:2209.14988}, 2022.

\bibitem[Qian et~al.(2023)Qian, Mai, Hamdi, Ren, Siarohin, Li, Lee, Skorokhodov, Wonka, Tulyakov, et~al.]{qian2023magic123}
Guocheng Qian, Jinjie Mai, Abdullah Hamdi, Jian Ren, Aliaksandr Siarohin, Bing Li, Hsin-Ying Lee, Ivan Skorokhodov, Peter Wonka, Sergey Tulyakov, et~al.
\newblock Magic123: One image to high-quality 3d object generation using both 2d and 3d diffusion priors.
\newblock \emph{arXiv preprint arXiv:2306.17843}, 2023.

\bibitem[Radford et~al.(2021)Radford, Kim, Hallacy, Ramesh, Goh, Agarwal, Sastry, Askell, Mishkin, Clark, et~al.]{radford2021learning}
Alec Radford, Jong~Wook Kim, Chris Hallacy, Aditya Ramesh, Gabriel Goh, Sandhini Agarwal, Girish Sastry, Amanda Askell, Pamela Mishkin, Jack Clark, et~al.
\newblock Learning transferable visual models from natural language supervision.
\newblock In \emph{International conference on machine learning}, pages 8748--8763. PMLR, 2021.

\bibitem[Raffel et~al.(2020)Raffel, Shazeer, Roberts, Lee, Narang, Matena, Zhou, Li, and Liu]{raffel2020exploring}
Colin Raffel, Noam Shazeer, Adam Roberts, Katherine Lee, Sharan Narang, Michael Matena, Yanqi Zhou, Wei Li, and Peter~J Liu.
\newblock Exploring the limits of transfer learning with a unified text-to-text transformer.
\newblock \emph{Journal of machine learning research}, 21\penalty0 (140):\penalty0 1--67, 2020.

\bibitem[Rajbhandari et~al.(2022)Rajbhandari, Li, Yao, Zhang, Aminabadi, Awan, Rasley, and He]{rajbhandari2022deepspeed}
Samyam Rajbhandari, Conglong Li, Zhewei Yao, Minjia Zhang, Reza~Yazdani Aminabadi, Ammar~Ahmad Awan, Jeff Rasley, and Yuxiong He.
\newblock Deepspeed-moe: Advancing mixture-of-experts inference and training to power next-generation ai scale.
\newblock In \emph{International conference on machine learning}, pages 18332--18346. PMLR, 2022.

\bibitem[Ramesh et~al.(2022)Ramesh, Dhariwal, Nichol, Chu, and Chen]{ramesh2022hierarchical}
Aditya Ramesh, Prafulla Dhariwal, Alex Nichol, Casey Chu, and Mark Chen.
\newblock Hierarchical text-conditional image generation with clip latents.
\newblock \emph{arXiv preprint arXiv:2204.06125}, 1\penalty0 (2):\penalty0 3, 2022.

\bibitem[Renggli et~al.(2022)Renggli, Pinto, Houlsby, Mustafa, Puigcerver, and Riquelme]{renggli2022learning}
Cedric Renggli, Andr{\'e}~Susano Pinto, Neil Houlsby, Basil Mustafa, Joan Puigcerver, and Carlos Riquelme.
\newblock Learning to merge tokens in vision transformers.
\newblock \emph{arXiv preprint arXiv:2202.12015}, 2022.

\bibitem[Riquelme et~al.(2021)Riquelme, Puigcerver, Mustafa, Neumann, Jenatton, Susano~Pinto, Keysers, and Houlsby]{riquelme2021scaling}
Carlos Riquelme, Joan Puigcerver, Basil Mustafa, Maxim Neumann, Rodolphe Jenatton, Andr{\'e} Susano~Pinto, Daniel Keysers, and Neil Houlsby.
\newblock Scaling vision with sparse mixture of experts.
\newblock \emph{Advances in Neural Information Processing Systems}, 34:\penalty0 8583--8595, 2021.

\bibitem[Rombach et~al.(2022)Rombach, Blattmann, Lorenz, Esser, and Ommer]{rombach2022high}
Robin Rombach, Andreas Blattmann, Dominik Lorenz, Patrick Esser, and Bj{\"o}rn Ommer.
\newblock High-resolution image synthesis with latent diffusion models.
\newblock In \emph{Proceedings of the IEEE/CVF conference on computer vision and pattern recognition}, pages 10684--10695, 2022.

\bibitem[Ronneberger et~al.(2015)Ronneberger, Fischer, and Brox]{ronneberger2015u}
Olaf Ronneberger, Philipp Fischer, and Thomas Brox.
\newblock U-net: Convolutional networks for biomedical image segmentation.
\newblock In \emph{Medical image computing and computer-assisted intervention--MICCAI 2015: 18th international conference, Munich, Germany, October 5-9, 2015, proceedings, part III 18}, pages 234--241. Springer, 2015.

\bibitem[Rosenbaum et~al.(2017)Rosenbaum, Klinger, and Riemer]{rosenbaum2017routing}
Clemens Rosenbaum, Tim Klinger, and Matthew Riemer.
\newblock Routing networks: Adaptive selection of non-linear functions for multi-task learning.
\newblock \emph{arXiv preprint arXiv:1711.01239}, 2017.

\bibitem[Ruiz et~al.(2021)Ruiz, Puigcerver, Mustafa, Neumann, Jenatton, Pinto, Keysers, and Houlsby]{ruiz2021scaling}
Carlos~Riquelme Ruiz, Joan Puigcerver, Basil Mustafa, Maxim Neumann, Rodolphe Jenatton, Andr{\'e}~Susano Pinto, Daniel Keysers, and Neil Houlsby.
\newblock Scaling vision with sparse mixture of experts.
\newblock In \emph{Advances in Neural Information Processing Systems}, 2021.

\bibitem[Salimans et~al.(2016)Salimans, Goodfellow, Zaremba, Cheung, Radford, and Chen]{salimans2016improved}
Tim Salimans, Ian Goodfellow, Wojciech Zaremba, Vicki Cheung, Alec Radford, and Xi~Chen.
\newblock Improved techniques for training gans.
\newblock \emph{Advances in neural information processing systems}, 29, 2016.

\bibitem[Sauer et~al.(2022)Sauer, Schwarz, and Geiger]{sauer2022stylegan}
Axel Sauer, Katja Schwarz, and Andreas Geiger.
\newblock Stylegan-xl: Scaling stylegan to large diverse datasets.
\newblock In \emph{ACM SIGGRAPH 2022 conference proceedings}, pages 1--10, 2022.

\bibitem[Shazeer et~al.(2017{\natexlab{a}})Shazeer, Mirhoseini, Maziarz, Davis, Le, Hinton, and Dean]{shazeer2017sparsely}
N~Shazeer, A~Mirhoseini, K~Maziarz, A~Davis, Q~Le, G~Hinton, and J~Dean.
\newblock The sparsely-gated mixture-of-experts layer.
\newblock \emph{Outrageously large neural networks}, 2017{\natexlab{a}}.

\bibitem[Shazeer et~al.(2017{\natexlab{b}})Shazeer, Mirhoseini, Maziarz, Davis, Le, Hinton, and Dean]{shazeer2017outrageously}
Noam Shazeer, Azalia Mirhoseini, Krzysztof Maziarz, Andy Davis, Quoc Le, Geoffrey Hinton, and Jeff Dean.
\newblock Outrageously large neural networks: The sparsely-gated mixture-of-experts layer.
\newblock \emph{arXiv preprint arXiv:1701.06538}, 2017{\natexlab{b}}.

\bibitem[Shen et~al.(2019)Shen, Ott, Auli, and Ranzato]{shen2019mixture}
Tianxiao Shen, Myle Ott, Michael Auli, and Marc’Aurelio Ranzato.
\newblock Mixture models for diverse machine translation: Tricks of the trade.
\newblock In \emph{International conference on machine learning}, pages 5719--5728. PMLR, 2019.

\bibitem[Singer et~al.(2022)Singer, Polyak, Hayes, Yin, An, Zhang, Hu, Yang, Ashual, Gafni, et~al.]{singer2022make}
Uriel Singer, Adam Polyak, Thomas Hayes, Xi~Yin, Jie An, Songyang Zhang, Qiyuan Hu, Harry Yang, Oron Ashual, Oran Gafni, et~al.
\newblock Make-a-video: Text-to-video generation without text-video data.
\newblock \emph{arXiv preprint arXiv:2209.14792}, 2022.

\bibitem[Sohl-Dickstein et~al.(2015)Sohl-Dickstein, Weiss, Maheswaranathan, and Ganguli]{sohl2015deep}
Jascha Sohl-Dickstein, Eric Weiss, Niru Maheswaranathan, and Surya Ganguli.
\newblock Deep unsupervised learning using nonequilibrium thermodynamics.
\newblock In \emph{International conference on machine learning}, pages 2256--2265. PMLR, 2015.

\bibitem[Song et~al.(2020)Song, Sohl-Dickstein, Kingma, Kumar, Ermon, and Poole]{song2020score}
Yang Song, Jascha Sohl-Dickstein, Diederik~P Kingma, Abhishek Kumar, Stefano Ermon, and Ben Poole.
\newblock Score-based generative modeling through stochastic differential equations.
\newblock \emph{arXiv preprint arXiv:2011.13456}, 2020.

\bibitem[Sun et~al.(2024)Sun, Jiang, Chen, Zhang, Peng, Luo, and Yuan]{sun2024autoregressive}
Peize Sun, Yi~Jiang, Shoufa Chen, Shilong Zhang, Bingyue Peng, Ping Luo, and Zehuan Yuan.
\newblock Autoregressive model beats diffusion: Llama for scalable image generation.
\newblock \emph{arXiv preprint arXiv:2406.06525}, 2024.

\bibitem[Tani and Nolfi(1999)]{tani1999learning}
Jun Tani and Stefano Nolfi.
\newblock Learning to perceive the world as articulated: an approach for hierarchical learning in sensory-motor systems.
\newblock \emph{Neural Networks}, 12\penalty0 (7-8):\penalty0 1131--1141, 1999.

\bibitem[Xin et~al.(2020)Xin, Tang, Lee, Yu, and Lin]{xin2020deebert}
Ji~Xin, Raphael Tang, Jaejun Lee, Yaoliang Yu, and Jimmy Lin.
\newblock Deebert: Dynamic early exiting for accelerating bert inference.
\newblock \emph{arXiv preprint arXiv:2004.12993}, 2020.

\bibitem[Yang et~al.(2019)Yang, Bender, Le, and Ngiam]{yang2019condconv}
Brandon Yang, Gabriel Bender, Quoc~V Le, and Jiquan Ngiam.
\newblock Condconv: Conditionally parameterized convolutions for efficient inference.
\newblock \emph{Advances in neural information processing systems}, 32, 2019.

\bibitem[Yang et~al.(2023)Yang, Zhang, Song, Hong, Xu, Zhao, Zhang, Cui, and Yang]{yang2023diffusion}
Ling Yang, Zhilong Zhang, Yang Song, Shenda Hong, Runsheng Xu, Yue Zhao, Wentao Zhang, Bin Cui, and Ming-Hsuan Yang.
\newblock Diffusion models: A comprehensive survey of methods and applications.
\newblock \emph{ACM Computing Surveys}, 56\penalty0 (4):\penalty0 1--39, 2023.

\bibitem[Yu et~al.(2018)Yu, Yang, Xu, Yang, and Huang]{yu2018slimmable}
Jiahui Yu, Linjie Yang, Ning Xu, Jianchao Yang, and Thomas Huang.
\newblock Slimmable neural networks.
\newblock \emph{arXiv preprint arXiv:1812.08928}, 2018.

\bibitem[Yuksel et~al.(2012)Yuksel, Wilson, and Gader]{yuksel2012twenty}
Seniha~Esen Yuksel, Joseph~N Wilson, and Paul~D Gader.
\newblock Twenty years of mixture of experts.
\newblock \emph{IEEE transactions on neural networks and learning systems}, 23\penalty0 (8):\penalty0 1177--1193, 2012.

\bibitem[Zeevi et~al.(1996)Zeevi, Meir, and Adler]{zeevi1996time}
Assaf Zeevi, Ron Meir, and Robert Adler.
\newblock Time series prediction using mixtures of experts.
\newblock \emph{Advances in neural information processing systems}, 9, 1996.

\bibitem[Zhang et~al.(2023)Zhang, Cai, Chen, Zhang, Zhang, Chen, Chang, Wang, and Liu]{zhang2023robust}
Yihua Zhang, Ruisi Cai, Tianlong Chen, Guanhua Zhang, Huan Zhang, Pin-Yu Chen, Shiyu Chang, Zhangyang Wang, and Sijia Liu.
\newblock Robust mixture-of-expert training for convolutional neural networks.
\newblock In \emph{Proceedings of the IEEE/CVF International Conference on Computer Vision}, pages 90--101, 2023.

\bibitem[Zoph et~al.(2022)Zoph, Bello, Kumar, Du, Huang, Dean, Shazeer, and Fedus]{zoph2022designing}
Barret Zoph, Irwan Bello, Sameer Kumar, Nan Du, Yanping Huang, Jeff Dean, Noam Shazeer, and William Fedus.
\newblock Designing effective sparse expert models.
\newblock \emph{arXiv preprint arXiv:2202.08906}, 2\penalty0 (3):\penalty0 17, 2022.

\end{thebibliography}

\end{document}